\documentclass{article}
\usepackage{spconf,amsmath,epsfig}

\usepackage[utf8]{inputenc} % allow utf-8 input
\usepackage[T1]{fontenc}    % use 8-bit T1 fonts
\usepackage{hyperref}       % hyperlinks
\usepackage{url}            % simple URL typesetting
\usepackage{booktabs}       % professional-quality tables
\usepackage{amsfonts}       % blackboard math symbols
\usepackage{nicefrac}       % compact symbols for 1/2, etc.
\usepackage{microtype}      % microtypography
\usepackage{adjustbox,lipsum} % to compress table size
\newcommand{\bftab}{\fontseries{b}\selectfont}
\usepackage{caption} % @sh
\title{Attention Based Semantic Segmentation on UAV Dataset For Natural Disaster Damage Assessment}
\name{Tashnim Chowdhury, Maryam Rahnemoonfar  \thanks{This work is partly supported by Microsoft AI for Earth grant. This paper has been accepted in the IGARSS 2021.} }
\address{Computer Vision and Remote Sensing Laboratory (Bina Lab)\\
    Department of Information Systems\\
    University of Maryland Baltimore County\\
	Baltimore, Maryland, USA}
\begin{document}
%\ninept
%
\maketitle
\begin{abstract}
The detrimental impacts of climate change include stronger and more destructive hurricanes happening all over the world. Identifying different damaged structures of an area including buildings and roads are vital since it helps the rescue team to plan their efforts to minimize the damage caused by a natural disaster. Semantic segmentation helps to identify different parts of an image. We implement a novel self-attention based semantic segmentation model on a high resolution UAV dataset and attain Mean IoU score of around $88\%$ on the test set. The result inspires to use self-attention schemes in natural disaster damage assessment which will save human lives and reduce economic losses.
\end{abstract}
\section{Introduction}
\label{sec:intro}

In recent time, the world has seen numerous natural disasters which have brought both personal injury and economic loss to several countries all over the world. A major step to both save valuable human lives and reduce financial loss is an accurate assessment of the damage inflicted by these events. A correct estimation of the damage helps to plan efficiently and allows the rescue team to allocate their efforts and aid properly.

With the advent of new sophisticated technologies capturing events of natural disaster has improved. Currently satellite imageries \cite{rudner2019multi3net, gupta2020rescuenet} have been used for proper assessment of these disaster events. The rescue team can use UAV (Unmanned Aerial Vehicle) to capture images of damaged properties and the whole affected area. Compared to satellite imagery, UAV imagery \cite{chowdhury2020comprehensive, rahnemoonfar2020comprehensive, rahnemoonfar2020floodnet} provides higher resolution which helps to understand the detailed damage level of the captured area.

Semantic segmentation is a core part of image understanding. Semantic segmentation is the task of assigning semantic labels to every pixels of an image. Besides traditional approach to tackle computer vision problems like semantic segmentation, deep convolutional neural networks  \cite{long2015fully} have achieved several revolutionary achievements in answering complex image understanding issues. Although many advanced segmentation works have been proposed in the recent years for popular urban datasets like Cityscapes \cite{cordts2016cityscapes} and PASCAL VOC \cite{mottaghi2014role} dataset, very few methods \cite{rudner2019multi3net, gupta2020rescuenet, zhu2020msnet, rahnemoonfar2018flooded} have been proposed and applied on natural disaster datasets.

Attention based models are able to capture long range dependencies and have been applied to several applications including computer vision field \cite{fu2019dual, huang2019ccnet}. Although self attention based models is gaining popularity in semantic segmentation of popular public non-disaster datasets like Cityscapes and PASCAL VOC due their higher performances, to the best of our knowledge no attention based model has been proposed for the semantic segmentation of disaster datasets.

\begin{figure}[t]
	\begin{center}
		\includegraphics[width=\linewidth]{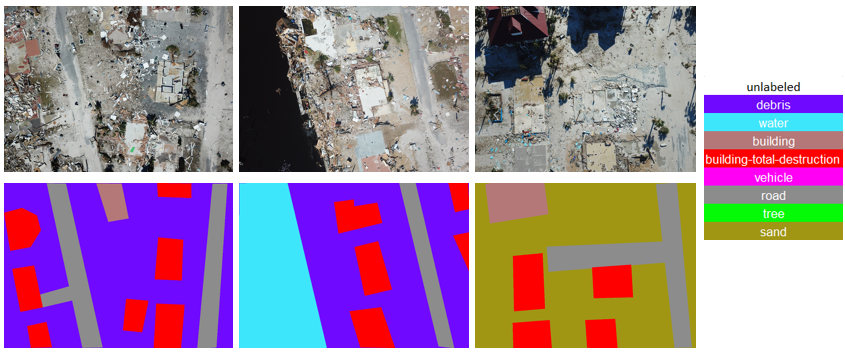}
	\end{center}
	\caption{Illustration  of  after disaster scenes  of  HRUD  dataset.  First  row  shows  original  image  and  the  second  row  shows  the corresponding annotations.}
	\label{fig:dataset-example}
\end{figure}

We propose a self attention method named ReDNet, a ResNet \cite{he2016deep} based Dual attention method, which aims to capture both spatial and semantic contribution of the features. We evaluate this model on High Resolution UAV Dataset (HRUD) \cite{chowdhury2020comprehensive, rahnemoonfar2020comprehensive}. Figure \ref{fig:dataset-example} shows sample images and respective colored annotated masks from HRUD. Most of the already proposed methods \cite{gupta2020rescuenet, rahnemoonfar2018flooded} have performed segmentation on flood water, damaged road, and buildings. But we comprehensively perform semantic segmentation on all of the components present in the images including debris, water, building, vehicle, road, tree, and sand. We also compare our method with three other state-of-art semantic segmentation models named ENet \cite{paszke2016enet}, DeepLabv3+ \cite{chen2018encoder}, and PSPNet \cite{zhao2017pyramid}.

\section{Related Works}
\label{sec:related-works}

\subsection{Semantic Segmentation}
Fully Convolutional Networks (FCNs) based methods have achieved excellent performance in semantic segmentation task. Popular state-of-art semantic segmentation methods can be classified into two sections: encoder-decoder based architecture \cite{paszke2016enet} and pooling based architecture \cite{zhao2017pyramid, chen2018encoder}. Encoder-decoder methods use middle and lower level features and generate local contextual map which creates sharp object boundaries. On the other hand, pooling based methods like PSPNet \cite{zhao2017pyramid} and DeepLabv3+ \cite{chen2018encoder} use pyramid pooling operations to create feature maps which are rich in global contextual information. Recently self-attention based methods \cite{huang2019ccnet, fu2019dual} have demonstrated remarkable performance in collecting global contextual dependencies and thus superior performance in segmentation.

\subsection{Natural Disaster Damage Assessment}
Recently several research works have been proposed addressing the damage assessment from natural disaster datasets. Rahnemoonfar \textit{et al.} implement a densely connected recurrent neural network in \cite{rahnemoonfar2018flooded} on UAV imageries for river segmentation. Rudner \textit{et al.} present a novel approach named Multi3Net in \cite{rudner2019multi3net} for segmentation of flooded buildings using satellite imageries. RescueNet is proposed by Gupta \textit{et al.} in \cite{gupta2020rescuenet} for joint building segmentation. Zhu \textit{et al.} propose an instance segmentation named MSNet in \cite{zhu2020msnet} on aerial videos to assess building damage.

In this paper, inspired by the recent performance of the self-attention based methods we propose a novel self-attention based network ReDNet and evaluate its performance on HRUD dataset \cite{chowdhury2020comprehensive, rahnemoonfar2020comprehensive}. To the best of our knowledge this is the first self-attention based method applied on any natural disaster dataset.

\section{Dataset}
\label{sec:dataset}

\begin{figure}[!htp]
	\begin{center}
		\includegraphics[width=0.8\linewidth]{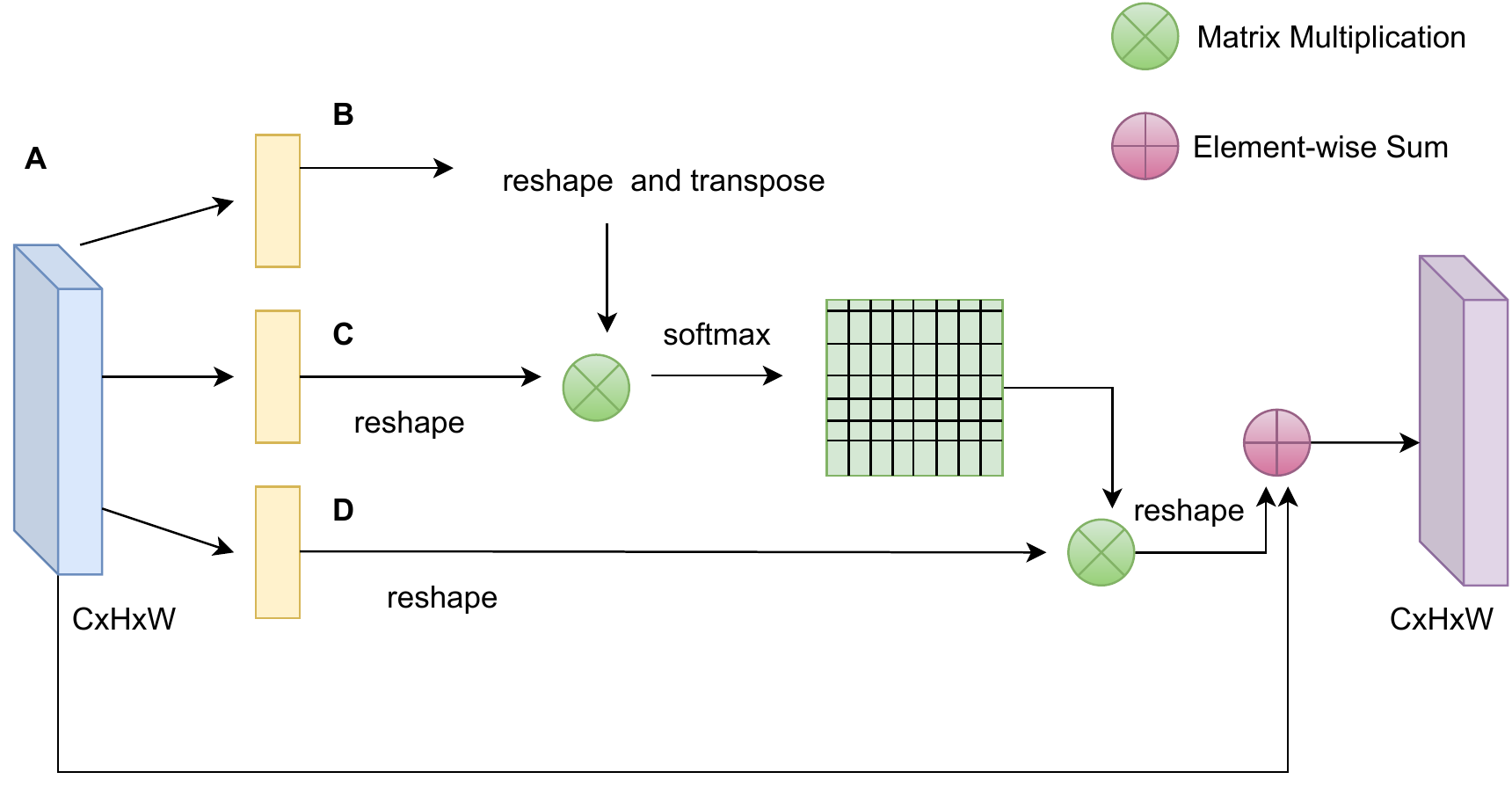}
	\end{center}
	\caption{The details of proposed position attention module.}
	\label{fig:pam}
\end{figure}

We use High Resolution UAV Dataset (HRUD) \cite{chowdhury2020comprehensive, rahnemoonfar2020comprehensive} where resolution of the captured images are 3000 $\times$ 4000. The dataset is taken from the Center for Robot-Assisted Search and Rescue open data repository (hrail.crasar.org) for small UAV imagery for disasters, specifically the Hurricane Michael event. The dataset is unique for two reasons. One is fidelity: it contains imagery from sUAV taken during the response phase by emergency responders, thus the data reflects what is the state of the practice and can be reasonable expected to be collected during a disaster. Second: it is the only known database of sUAV imagery for disasters. Note that there are other existing databases of imagery from unmanned and manned aerial assets collected during disasters, such as National Guard Predators or Civil Air Patrol, but those are larger, fixed-wing assets that operate above the 400 feet AGL limitation of sUAV. The HRUD dataset consists of imagery taken from 80 flights conducted between October 11-14, 2018,  on behalf of the Florida State Emergency Response Team at Mexico Bearch and other directly impacted areas. All flights were flown at 200 feet AGL, as compared to manned assets which normally fly at 500 feet AGL or higher. The majority of daylight images were taken with DJI Mavic Pro quadcopters, though two sets of video were taken with Parrot Disco fixed-wing sUAV and one set at night with a DJI Inspire and thermal camera.

\begin{figure}[!htp]
	\begin{center}
		\includegraphics[width=\linewidth]{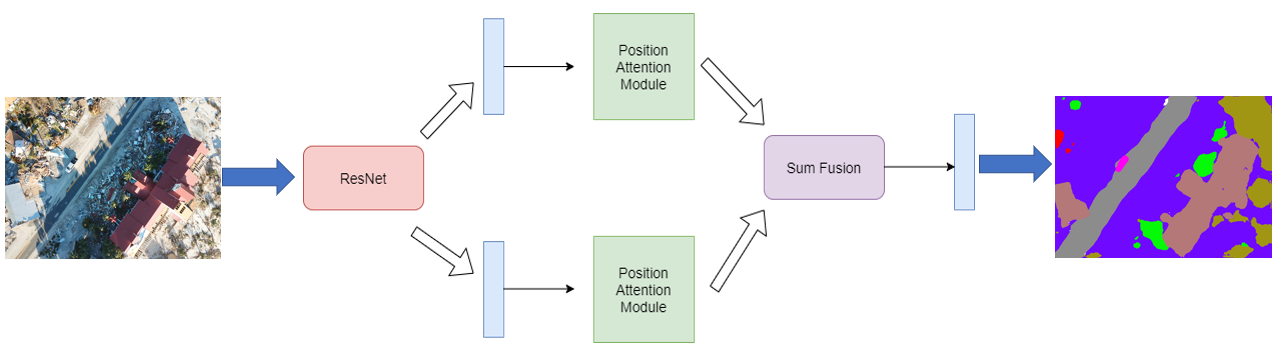}
	\end{center}
	\caption{An overview of our ReDNet architecture.}
	\label{fig:rednet-arch}
\end{figure}

The images include debris, water, building, vehicle, road, tree, and sand. The images are collected after hurricane Michael. In total 1973 images have been annotated for semantic segmentation. The buildings are annotated into four classes based on their damage levels. The classes are no damage, medium damage, major damage, and total destruction. For the experiments in this paper, we merge no damage, medium damage, and major damage into ``Building Not Totally Destroyed'' and total destruction is considered as ``Building Totally Destroyed''.

\begin{table*}[!htp]
	\centering
	\caption{Per-class results on HRUD testing set.}\label{table:michael-test-perfor-table}
	\begin{adjustbox}{width=\textwidth}
	\begin{tabular}{|l c c c c c c c c c| c}
	\hline
		Method & Debris & Water &  \begin{tabular}{@{}c@{}}Building Not \\  Totally Destroyed\end{tabular} & \begin{tabular}{@{}c@{}}Building \\  Totally Destroyed\end{tabular} & Vehicle & Road & Tree & Sand & mIoU\\\hline
		\hline
		ENet\cite{paszke2016enet} & 45.97 & 75.84 & 66.16 & 39.52 & 36.74 & 61.19 & 71.64 & 61.77 & 54.15\\
		DeepLabv3+\cite{chen2018encoder} & 65.8 & 85.8 & 84.5 & 57.3 & 51.3 & 73.3 & 75.9 & 77.4 & 69.67\\
		PSPNet\cite{zhao2017pyramid} & 88.76 & 67.98 & 85.75 & 80.51 & \bftab 65.83 & 82.81 & 94.53 & 76.04 & 80.27\\ 
		ReDNet(Our method) & \bftab 92.66 &  \bftab 95.80 & \bftab 92.16 & \bftab 83.01 & 57.3 & \bftab 89.78 & \bftab 96.41 & \bftab 96.46 & \bftab 87.95\\ \hline
	\end{tabular}
	\end{adjustbox}
\end{table*}

\begin{figure*}[!htp]
	\begin{center}
		\includegraphics[width=0.7\linewidth]{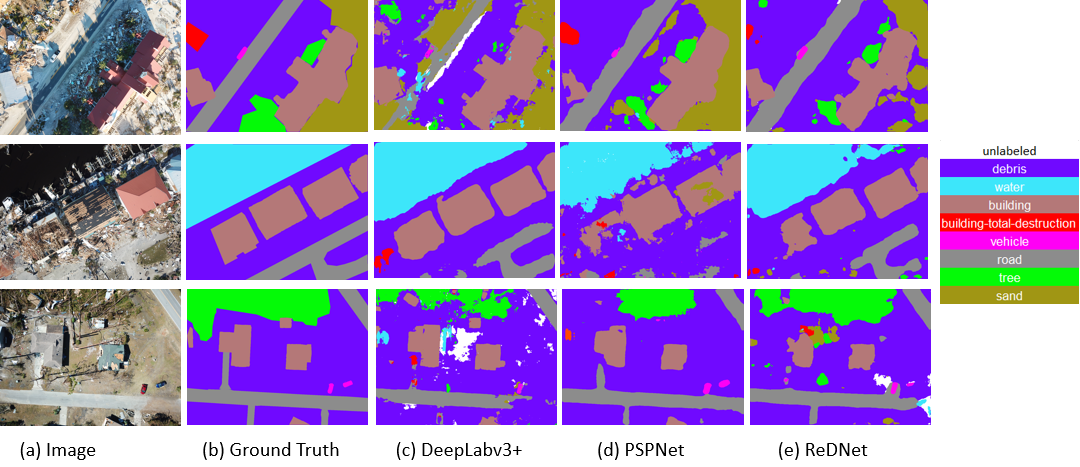}
	\end{center}
	\caption{Visual comparison on HRUD test set.}
	\label{fig:vis-compare-models-all-cls}
\end{figure*}

\section{Method}
\label{sec:method}

\subsection{PAM: Position Attention Module}
Convolution operations lead to local receptive field. Lack of global context results in intra-class inconsistency and hurts the recognition accuracy \cite{fu2019dual, huang2019ccnet}. We use ResNet-101 as base recognition model. Different stages of the ResNet-101 has different recognition capability. Lower stages show encoding of fine spatial information but poor semantic consistency due to smaller receptive field view. On other hand, higher stages shows good semantic consistency due to larger receptive field but poor encoding of spatial information. Keeping this in mind, we propose a novel self-attention segmentation network ReDNet which combines two position attention modules by extracting features from both lower and higher stages of ResNet-101.

Position attention module is shown in Figure \ref{fig:pam}. Given a local feature map $A$ with dimension $C \times H \times W$ we generate three new feature maps using a convolution layer with similar shape to given feature map $A$. Matrix multiplications is performed on transpose of $B$ and $C$. Then using a softmax layer spatial attention map $S$ is calculated using the following formula.

\begin{equation}
    s_{ji} = \frac{exp(B_i \cdot C_j)}{\sum_{i=1}^N exp(B_i \cdot C_j)}
\end{equation}

Then matrix multiplication is performed between transpose of $S$ and $D$. The output from the matrix multiplication is multiplied with $\alpha$. Finally element-wise sum is performed with feature $A$ to generate final output with shape $C \times H \times W$.

\subsection{ReDNet}

ReDNet architecture consists of two position attention modules (PAMs) which are placed in parallel. One PAM takes output features from layer 2 of ResNet-101 as input while the other one takes output of layer 3 of ResNet-101 as input. Their attention maps are added together using element-wise sum operation. ReDNet architecture is shown in Figure \ref{fig:rednet-arch}.

\section{Experiments} 
\label{sec:experiment}

\subsection{Implementation Details}
PyTorch has been used for implementation of segmentation networks. As hardware we use NVIDIA GeForce RTX 2080 Ti GPU and Intel Core i9 CPU. We use ``poly'' learning rate with base learning rate 0.0001. Momentum, weight decay, and power are set to 0.9, 0.0001, and 0.9 respectively. For augmentation we use random shuffling, scaling, flipping, and random rotation which help models to avoid overfitting. During training, we resize the images to 713 $\times$ 713 since large crop size is useful for the high resolution images. For semantic segmentation evaluation metric we use mean IoU.

\section{Results and Discussion}
\label{sec:res-dis}

\par HRUD is a very challenging dataset due to its variable sized classes along with similar textures among different classes. Debris makes a great impact on segmentation performances of the evaluated network models. Similar textures of debris, sand, and building with total destruction damage are very difficult to differentiate from segmentation performances' point of view.

\par Evaluations of our method ReDNet on HRUD test set is presented in Table \ref{table:michael-test-perfor-table} and compared with three other state-of-art segmentation models, ENet \cite{paszke2016enet}, DeepLabv3+ \cite{chen2018encoder}, and PSPNet \cite{zhao2017pyramid}. ReDNet achieves best results in all classes with mean IoU of $87.95\%$. Superior performance of ReDNet proves that global feature map generated using proposed self-attention method improves the segmentation compared to extracted local context map by ENet, and global feature maps by PSPNet and DeepLabv3+. The visualization of the segmentation can be seen in Figure \ref{fig:vis-compare-models-all-cls}.

\section{Conclusion}
\label{conclusion}

In this paper, we propose and evaluate a novel self-attention segmentation model ReDNet on a new high resolution natural disaster dataset named HRUD. We discuss the challenges of semantic segmentation on this dataset along with the excellent performance of the proposed model on it. We hope this paper will facilitate future research on natural disaster damage assessment and will help achieving less human and economic losses with efficient natural disaster management.

% Below is an example of how to insert images. Delete the ``\vspace'' line,
% uncomment the preceding line ``\centerline...'' and replace ``imageX.ps''
% with a suitable PostScript file name.
% -------------------------------------------------------------------------

% To start a new column (but not a new page) and help balance the last-page
% column length use \vfill\pagebreak.
% -------------------------------------------------------------------------
%\vfill
%\pagebreak

% References should be produced using the bibtex program from suitable
% BiBTeX files (here: strings, refs, manuals). The IEEEbib.bst bibliography
% style file from IEEE produces unsorted bibliography list.
% -------------------------------------------------------------------------
\bibliographystyle{IEEEbib}
%\bibliography{strings,refs}
\bibliography{main}

\end{document}